\documentclass[10pt]{article} % For LaTeX2e
\usepackage[accepted]{tmlr}
\usepackage{times}
\usepackage{epsfig}
\usepackage{graphicx}
\usepackage{amsmath}
\usepackage{amssymb}
\usepackage{multirow}
\usepackage{tabu}
\usepackage{url}
\usepackage{xcolor}
\usepackage{xspace}
\usepackage{verbatim} % for comment
\usepackage{subcaption}

\usepackage[pagebackref=true,breaklinks=true,letterpaper=true,colorlinks,bookmarks=false]{hyperref}
\newcommand{\tablestyle}[2]{\setlength{\tabcolsep}{#1}\renewcommand{\arraystretch}{#2}\centering\footnotesize}

\newcommand{\gray}[1]{\textcolor{gray}{{#1}}}

\newcommand{\white}[1]{\color[HTML]{FFFFFF}{{#1}}}

\newcommand{\ours}{MaMMUT\xspace}

\makeatletter
\newcommand{\thickhline}{%
    \noalign {\ifnum 0=`}\fi \hrule height 1pt
    \futurelet \reserved@a \@xhline
}

\usepackage[pagebackref=true,breaklinks=true,letterpaper=true,colorlinks,bookmarks=false]{hyperref}

\title{MaMMUT: A Simple Architecture for Joint Learning for MultiModal Tasks}
\author{
\parbox{\linewidth}{{\centering
Weicheng Kuo\footnotemark[1]\hspace{.4cm}
AJ Piergiovanni\footnotemark[1]\hspace{.4cm}
Dahun Kim\footnotemark[2]\hspace{.4cm}
Xiyang Luo\footnotemark[2]\hspace{.4cm}
Ben Cain\hspace{.4cm}
Wei Li\hspace{.4cm}
Abhijit Ogale\hspace{.4cm}
Luowei Zhou\hspace{.4cm}
Andrew Dai\hspace{.4cm}
Zhifeng Chen\hspace{.4cm}
Claire Cui\hspace{.4cm}
Anelia Angelova
\\
}
}
\parbox{\linewidth}{\centering \vspace{0.2cm}
Google Research\thanks{Correspondence to \texttt{\small{\{weicheng, ajpiergi, anelia\}}@google.com.}}
}
\vspace{0.5cm}
}
\vspace{-1cm}
\def\month{08}  % Insert correct month for camera-ready version
\def\year{2023} % Insert correct year for camera-ready version
\def\openreview{\url{https://openreview.net/forum?id=FqOG4osY7C}} % Insert correct link to OpenReview for camera-ready version

\begin{document}

\maketitle

%%%%%%%%% ABSTRACT
\vspace{-0.5cm}
\begin{abstract}
\vspace{-0.2cm}

 The development of language models have moved from encoder-decoder to decoder-only designs. In addition, we observe that the two most popular multimodal tasks, the generative and contrastive tasks, are nontrivial to accommodate in one architecture, and further need adaptations for downstream tasks. We propose a novel paradigm of training with a decoder-only model for multimodal tasks, which is surprisingly effective in jointly learning of these disparate vision-language tasks. This is done with a simple model, called MaMMUT. It consists of a single vision encoder and a text decoder, and is able to accommodate contrastive and generative learning by a novel two-pass approach on the text decoder. We demonstrate that joint learning of these diverse objectives is simple, effective, and maximizes the weight-sharing of the model across these tasks. Furthermore, the same architecture enables straightforward extensions to open-vocabulary object detection and video-language tasks. The model tackles a diverse range of tasks, while being modest in capacity. Our model achieves the state of the art on image-text and text-image retrieval, video question answering and open-vocabulary detection tasks, outperforming much larger and more extensively trained foundational models. It shows very competitive results on VQA and Video Captioning, especially considering its capacity. Ablations confirm the flexibility and advantages of our approach.

\end{abstract}
% \thanks{$*$, $\dagger$ : Equal contributions.}

\vspace{-5mm}
\section{Introduction}
\vspace{-2mm}
% \textbf{Outline v1}:
% \begin{itemize}
%  \item Vision and language tasks have exhibit many similarity and the recent propagation of transformers from language to vision is yet another solid evidence.
%  \item Language models have moved towards decoder only design as opposed to encoder-decoder. However, most VL systems still rely on both encoder and decoder on the language side, which motivates us to explore the potential of decoder-only language models.
%  \item reconciling contrastive and generative training is challenging and is done step-wise or with encoder-decoder
%  \item We propose an alternative design to encoder decoder models by flexibly injecting the vision knowledge at any layer of the decoder-only LM via cross attention. This supports both contrastive and generative vision-language tasks with fully shared weights other than the cross-attention module.
%  \item Experiments show that our design achieve SOTA or very competitive accuracy across a suite of VL tasks.
% \end{itemize}

Vision-language learning has become critical in improving both  visual-understanding and multimodal vision-language tasks.
%On one hand, training on automatically, on the other training is very easy %
%
%
%%
%
%One of the main advantages of image-text learning is the abundance of cheaply obtained, and often noisy, image-language examples pairs. 
Large foundational vision-language models,  which are designed to be extended to multiple downstream tasks, follow two main training strategies, typically exemplified by disjoint architectures. %accomplished by simple and well understood losses, 
Some vision-language pre-training approaches apply a contrastive loss, in a dual-encoder style architecture, e.g. CLIP, Align, Florence~\citep{radford2021clip,align,yuan2021florence}. 
Contrastive training has been shown to produce strong backbones, which lead to successful image understanding and cross-modal retrieval tasks, e.g. image-to-text or text-to-image retrieval. 

Alternatively, the autoregressive and masked token modeling objectives, well known from language modeling, are very popular with vision-language models for text generation. They are often referred to as split-captioning objectives. 
  The split-captioning training is typically beneficial to text-generative tasks e.g. VQA~\citep{agrawal2015vqa}. 
 
The most common architectures used in these scenarios are the encoder-decoder ones, which use a separate vision and text encoders, or a joint vision-text encoder, before a joint decoder, applying decoding %, extending the encoding and decoding 
losses from language learning~\citep{pmlr-v139-cho21a,pali,piergiovanni2022answer,wang2021simvlm,omniVL}. Architectures with cross-attention over frozen or partly frozen language models have also been popular~\citep{flamingo}.

Combining these two types of architectures and loss functions has proven to be challenging, with recent approaches such as  Align-Before-Fuse (ALBEF) and CoCa~\citep{albef,yu2022coca,videococa} requiring multiple components or training stages, and special recipe to accommodate video tasks~\citep{videococa}. Simultaneously, many pure language-only models have adopted simple decoder-only architectures with great success~\citep{liu2018generating,gpt3,glam} and with the added benefit of significant parameter reduction~\citep{liu2018generating}.

We here propose a simple approach to unify contrastive learning, localization aware, and autoregressive captioning pretraining, by using a single language decoder and an image encoder.
%We inject vision features from the encoder by cross-attention. 
Our formulation is more general and allows maximal weight-sharing and parameter efficiency between the contrastive and generative tasks. % in the text decoder.
%Decoder-only models offer clear advantages in performance and parameter savings for language learning, however, one challenge for multi-modal tasks is reconciling the unconditional sequence-level representation learning needed for contrastive learning with the token-conditioned next-token prediction.    
To address the  challenge of reconciling the unconditional sequence-level representation learning needed for contrastive learning with the token-conditioned next-token prediction,  we propose a two-pass learning strategy using the text decoder. In one pass of the training, we utilize cross attention and causal masking to learn the caption generation task, where the text features can attend to the image features and predict the tokens in sequence; in the other pass we disable the cross-attention and causal masking, which learns the contrastive task without visibility into the image features. We further modify the contrastive training objective to be localization-aware, further equipping the model for object detection tasks. With this training strategy our model can address a diverse range of tasks, e.g. retrieval, text generative or detection.
This provides a simpler alternative to previous approaches~\citep{ albef,yu2022coca,singh2022flava,yuan2021florence}, where our model architecture is more compact and shared more broadly across tasks in the model. % and further allows for language-only decoder reuse.
Furthermore, our model allows us to apply this architecture to video by a seamless adaptation based on the TubeViT approach~\citep{piergiovanni2022tubevit}. This allows us to address video-text tasks, such as Video Question Answering (VideoQA) and Video Captioning successfully,
outperforming prior large image-text or video foundatinoal models, such as 80B Flamingo, by pre-training on image-and-text dataset only.  Furthermore, we extend the approach to leverage the pre-trained model for open-vocabulary detection, demonstrating the localization capabilities of the model.
We call the model \ours.

 \begin{figure*}
    \centering
    \vspace{-0.7cm}
   \includegraphics[width=0.95\linewidth]{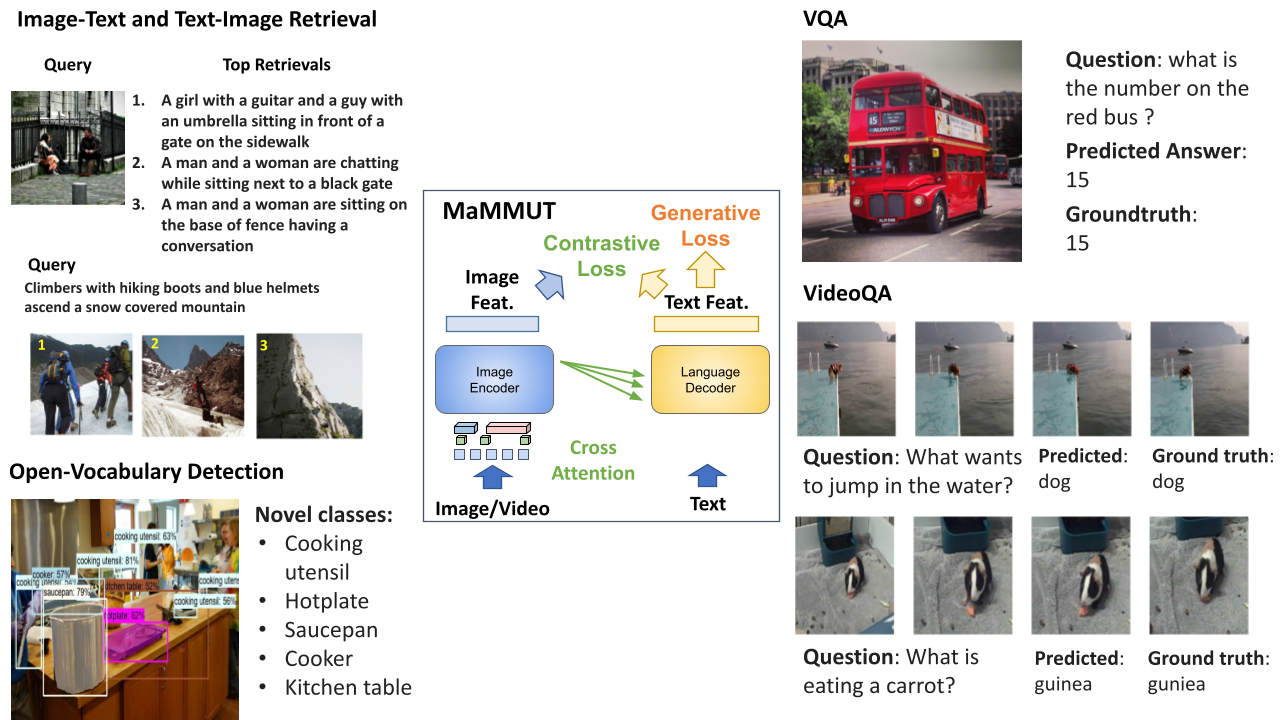}
    \vspace{-0.1cm}
    \caption{The \ours model is a simple vision-encoder and text-decoder architecture, which serves as foundational model for both image-language and video-language tasks. Despite its relatively small size, the model outperforms SOTA on many diverse tasks. Example results on  Image-Text/Text-Image retrieval, Visual Question Answering (VQA), Open-vocabulary detection, Video Question Answering (VideoQA) are shown. % and Video Question Answering (VideoQA). %TODO (re-do videoQA visualizations)
    }
    \label{fig:teaser}
    \vspace{-0.5cm}
\end{figure*}

Our model is simple, but at the same time represents a union of tasks that other models find it challenging to put together, or need more specialized adaptations for. 
%The above mentioned architecture have lead to a number of powerful models, which when scaled.
In comparison, many of the prior approaches, despite reporting results on 10, 20 or more tasks or training bigger models ~\citep{pali,singh2022flava,flamingo,yuan2021florence}, accommodate fewer categories of tasks than our model does. One of the largest models, PaLI~\citep{pali} addresses only classification, VQA and image captioning task categories, but is not designed to handle retrieval, detection, %, which are better suited to the dual-encoder architectures~\citep{radford2021clip,align}, 
or video tasks. 
 GIT~\citep{wang2022git} extends the model to videos, % processing a single frame at a time, 
 but does not handle retrieval or detection.
FLAVA~\citep{singh2022flava}, %or object localization tasks. 
Florence~\citep{yuan2021florence}, ALBEF~\citep{albef}, BLIP~\citep{li2022blip} which are based on constrastive learning,  accommodate more tasks, %significant downstream adaptations are needed for VQA-like tasks.% are needed to extend the task offering. 
but do not perform video-language or object localization tasks, or need significant downstream adaptations to do so~\citep{yuan2021florence}.
Flamingo~\citep{flamingo} %, which incorporates video and image tasks by design, 
is not able to do retrieval or object detection.
With \ours, we address all these tasks.

Experiments on image-text, text-image retrieval, VideoQA and Open-Vocabulary Object Detection show above-SOTA performances, outperforming much larger models. \ours shows very competitive performance in Video Captioning and Visual Question Answering (VQA), which is done in the challenging open-ended generation setting. We note that our pre-training is done on image-text noisy pairs only and does not include video or labeled object localization information or supervised labels. Ablation studies confirm the challenges and benefits of developing a simple and flexible model to unify multiple tasks.

Our contributions are: 
(i) a simple, compact and extensible foundational vision-language model which addresses multiple diverse multimodal tasks, such as image/text retrieval, Visual Question Answering, Open-Vocabulary Detection, Video Question Answering and Video Captioning;
(ii) a novel two-pass learning method which trains jointly reconciling the unconditional sequence-level representation learning needed for contrastive learning with the token-conditioned captioning-like learning, using a single shared decoder-only model;
(iii) a seamless extension to video tasks with no additional pre-training needed, creating a powerful video-language model.

%\textbf{Outline}:
%\begin{itemize}
    %\item Vision and language pretraining is crucial to multimodal understanding. VL pretraining evolves from supervised learning to captioning and contrastive learning.
 %   \item Unifying multiple aspects of pre-training: contrastive, masking/next-token prediction and localization aware contributes to 
    %\item Combining contrastive learning and captioning pretraining enables a broad suite of both discriminative and generative tasks from zero-shot to fine-tuning settings.
    %\item We propose a simple approach to unify contrastive learning, localization aware and next-token captioning pretraining by using a single language decoder and inject vision features by cross-attention. Our formulation is more general and allows maximal weight-sharing and parameter efficiency between the contrastive and generative tasks in the text decoder.
    %\item We train generative and contrastive tasks jointly in one end-to-end stage and see promising multitask benefits.
    % \item VL pre-training is essential to building strong visual backbones; we show improved performance on vision-only downstream tasks. At the same time language data, intertwined during VL pretraining is crucial to enable tasks in the wild such as open-set object detection, which are not possible with detection datasets only.
%    \item Experiments on retrieval, VQA, captioning, and OVD show the effectiveness of our approach.
%\end{itemize}

%---------
\vspace{-2mm}
\section{Related Work}
\vspace{-2mm}
%\begin{itemize}
Vision and language pretraining has gained considerable popularity, where many successful and wide-ranging approaches have been created.
Following token masking or next-token prediction losses used in text modeling, image-language pre-training methods extended these ideas for image and text inputs, where a language modeling loss is applied to a model which considers image inputs   ~\citep{wang2022git,pali,wang2021simvlm,wang2022image,chen2020uniter,UnifiedVLP,VinVL,tan2019lxmert,li2020oscar,vilbert2020,su2020vlbert,meter}. %. That is, masking and next token prediction are applied to text inputs and outputs
This technique enables text-generative vision-and-language tasks, such as VQA, Image Captioning or classification. % but have also been applied to other downstream tasks such as classification, detection. 
Extensions to these approaches, where masking is done in the image space~\citep{bao2021beit}, have also been considered. 
Some of the above-mentioned approaches also feature similar architecture as ours, with a simple decoder e.g. GIT~\citep{wang2022git}, but the tasks addressed are limited to the text generative ones.

Contrastive vision and language pretraining has been popularized by the CLIP and Align models~\citep{radford2021clip,align}, demonstrating that contrastive learning can produce powerful embedding for many downstream applications. While the contrastive loss has been widely applicable, e.g. in self-supervised learning~\citep{chen2020simple}, the appeal of these methods is the ability to learn across two modalities and produce high quality embeddings from large amounts of noisy image-language pairs which are harvested automatically from the web. Many generative pretraining approaches have also subsequently leveraged such noisy and easily obtainable datasets for vision-language pre-training purposes.  
Contrastive vision-language models are typically two-tower models, where nontrivial modifications are needed for downstream tasks~\citep{yuan2021florence}.
%\item Combining contrastive and generative VLP.
  
Several prior works proposed approaches to combine contrastive and generative vision-language pre-training, on the premise of two-tower models and cross-attention or cross-modal masking to align the modalities~\citep{albef,singh2022flava,li2022blip,Li2023BLIP2BL,yu2022coca}. ALBEF~\citep{albef} applies a contrastive loss to an image and text encoder models and adds a decoder for generative tasks. BLIP-2~\citep{Li2023BLIP2BL} leverages an off-the-shelf frozen image encoder and large language model for generative learning. Similar to ALBEF~\citep{albef}, CoCa~\citep{yu2022coca} uses a language generation rather than masked language modeling objective. Furthermore, a second decoder uses cross-attention to connect image representations with text to generate the output text. In these approaches there is a distinction of unimodal vs multimodal text model, and contrastive loss is only applied to the unimodal part. Compared to CoCa, our approach differs in a few aspects: ours is a fully shared text model enabled by the proposed two-pass learning paradigm; it achieves state-of-the-art performance on video and open-vocabulary detection tasks which are not easily derived from language generation; has affordable computational cost for reproducibility; in terms of zero-shot performance, it has state-of-the-art zero-shot image-text retrieval. Other approaches~\citep{wang2022unifying,UnifiedIO} unify understanding and generation tasks in the same sequence-to-sequence framework, spanning a larger number of tasks, including tasks, such as, referring expressions~\citep{refcoco}. BEiT-3~\citep{bao2021beit} learns unimodal visual representation from image token reconstruction and finetunes the backbone for downstream tasks.
%, propose to start from the same core model, but re-configure it according to the type of task.
%However this approach requires significant changes of the model depending on the task.

 %Video-Language models are also becoming popular, where 
 A number of video foundational models have been proposed~\citep{merlot,fu2021violet,flamingo,internvideo,omniVL,VINDLU,UniVL}. The Flamingo model~\citep{flamingo} extends a large frozen language model, with image and video inputs to deliver impressive results. % over a range of vision-language tasks.
 VIOLET~\citep{fu2021violet} uses masked language and masked-video modeling for joint video-text learning. 
 Other approaches extend a pre-trained image-language model, adapting it to video~\citep{wang2022git,videococa,dynpretr}, where a common approach is to just accumulate features from individual frames of the video.
 Our model is mostly aligned to the latter class of models, however, we instead directly process the spatio-temporal information of the video and do not need to process the model by a single frame at a time. Our approach is more advantageous, as much of the temporal information is lost in the sinlge-frame approaches. 
 Florence~\citep{yuan2021florence} does an adaptation with 3D kernels to the SWIN transformer~\citep{swin}, and OmniVL~\citep{omniVL} uses the Timesformer model~\citep{timesformer}, which preserve the temporal information, however the adaptation is generally more complex.
 Combining contrastive and captioning losses in video is also a popular technique. For example, InternVideo~\citep{internvideo} proposed a combination of a masked video encoder-decoder and a cross-modal constrastive dual encoder in a video foundational model. 
 In contrast, % our approach is much simpler.
 while we perform only a light fine-tuning over a image-language pre-trained model, our approach outperforms the above-mentioned more  sophisticated and better-trained video models. 

 Object detection is available in some  foundation models, for example VinVL~\citep{VinVL} pre-train the vision-language model in order to detect objects and attributes. Florence~\citep{yuan2021florence}, adapt vision-language pre-trained models to object detection tasks. At the same time, Open-Vocabulary Detection, which aims at detecting novel categories of objects, is often overlooked by pre-trained vision-language models, where previous approaches focused on detecting objects in a set of pre-defined classes. VILD~\citep{gu2022openvocabulary}, 
 %and F-VLM~\citep{kuo2022fvlm} 
 demonstrate that large vision-language models are advantageous for detecting novel object categories. Inspired by this work, while taking a different approach, we show that open-vocabulary detection is an easy extension to our model with strong performance. 
%\end{itemize}
%---------
\begin{figure*}
    \centering
       \vspace{-0.7cm}
    \includegraphics[width=0.95\linewidth]{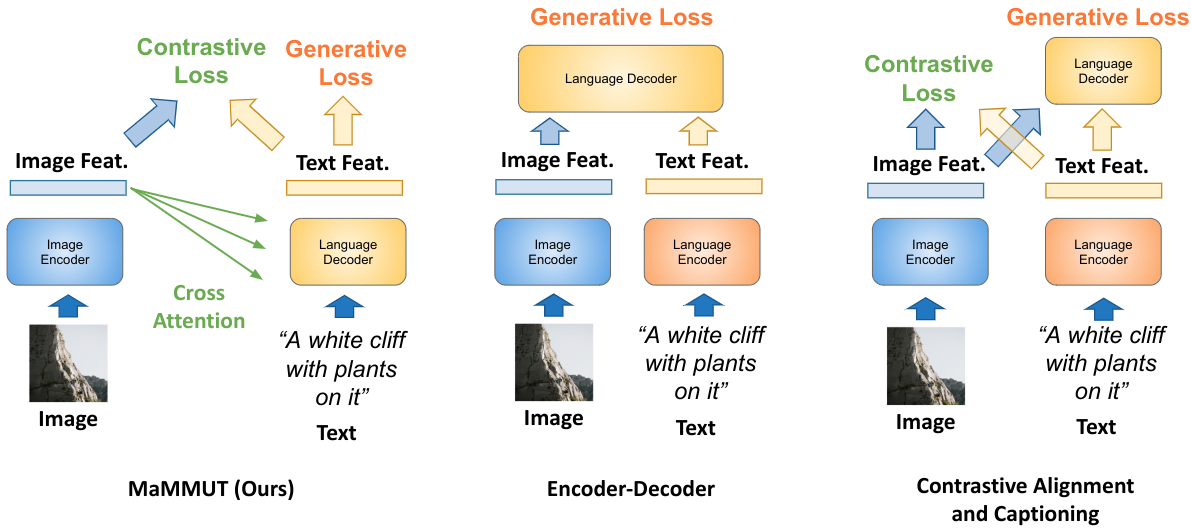}
     \vspace{-0.2cm}
    \caption{\ours model architecture with an image encoder and text decoder (left), compared to others. Many encoder-decoder architectures (center) cannot handle the contrastive objective, for example~\citep{pmlr-v139-cho21a,pali}. Approaches to combine contrastive and captioning (right), e.g. Align-Before-Fuse (ALBEF)~\citep{albef} or CoCa~\citep{yu2022coca} develop more complex models and are hard to extend to video inputs or localization tasks.  Our architecture is simpler than previous approaches and is able to accommodate more tasks.}
    \label{fig:main_arch}
    \vspace{-0.4cm}
\end{figure*}

\vspace{-0.1cm}
\section{Method}
\vspace{-0.3cm}
%We start by presenting the preliminaries on three families of modeling approaches that use natural language supervision. These include dual-encoder contrastive learning, encoder-decoder image captioning, and joint contrastive and captioning models. 
We introduce our model, called \ours, which offers a simple architecture consisting of a single image-encoder and a text-decoder (Figure~\ref{fig:main_arch}, left).
Our model combines the strengths of contrastive learning and autoregressive next-token prediction in a more flexible architecture than existing works. For example, the model is able to train and perform inference for those tasks with the same shared set of weights. 
Furthermore, one of the advantages of this simple model is that it allows simple extensions to video where the exact same architecture is able to consume directly video features. This is in contrast to prior work which adapt to video by processing individual frames or by more complex mechanisms. Other important downstream tasks, such as open-vocabulary detection, can also be easily added to this model. % as shown below.
%We also describe how \ours models can efficiently adapt to new tasks by zero-shot transfer or full finetuning.

% \begin{figure}
%     \centering
%     \includegraphics[width=1.0\linewidth]{MaMMUT/mammut_arch_v2.pdf}
%     \caption{\ours model architecture diagram V2.}
%     \label{fig:main_arch_v2}
% \end{figure}

%---------
\vspace{-0.2cm}
\subsection{\ours Architecture}
\vspace{-0.1cm}
\ours is an intuitive and simple architecture for multimodal tasks, which consists of a single vision-encoder and a single text-decoder (see left of Figure~\ref{fig:main_arch}). %Similar to prior vision-language architectures, 
We encode the images into latent representation using a neural network encoder (such as ViT~\citep{dosovitskiy2020image}). The texts are both encoded and decoded with a Transformer-based decoder-only model.
The \ours architecture utilizes cross attention to fuse visual representation with text features anywhere in the decoder layers.  
 This allows the whole decoder to produce both unimodal text representations (needed for a forward pass for contrastive learning) and multimodal text representation (needed to fuse the visual features with the text ones for vision+text tasks). This is done by a two pass joint training and we notably use a single combined training objective (shown later in Equation~\ref{eqn:combine}). Below we delve deeper into different aspects of the \ours model.

\begin{figure}
    \centering
     \vspace{-0.7cm}
    \includegraphics[width=0.7\linewidth]{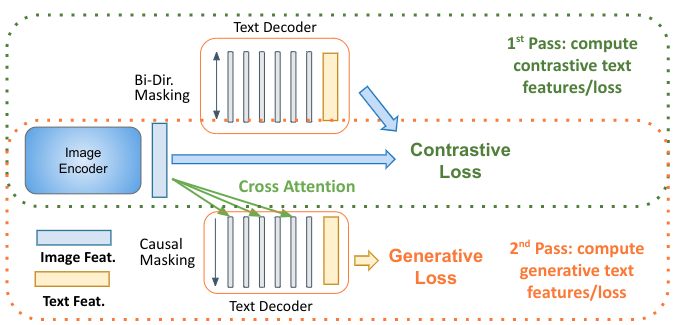}
     \vspace{-0.2cm}
    \caption{\ours two-pass learning. We compute the image features via the image encoder. Then we compute the contrastive text features and loss by applying a non-causal masking to the language decoder, which makes it effectively a text encoder (top). Finally we compute the generative text features and loss by applying causal masking and cross attention with the image features (bottom). The decoder is visualized twice for clarity, its weights are fully shared.}
    \vspace{-0.4cm}
    \label{fig:two-pass}
\end{figure}

\vspace{-0.1cm}
%\paragraph{Decoder-only for Contrastive and Captioning Tasks.}\quad 
\paragraph{Decoder-only Two-Pass learning.}\quad 
The main challenge to unify contrastive learning and next-token prediction is to unify the text representation, because the contrastive learning uses unconditional sequence-level representation, whereas captioning optimizes the likelihood of each token conditioned on the previous tokens. % (see Equation~\ref{eqn:gen}). 

We propose a two-pass approach to jointly learn the two types of text representations by the same model. During the first pass, to learn the contrastive task, we enable bi-directional masking within the decoder. The text features should not see the image features (which characterizes dual-encoder contrastive learner), but can attend to all tokens at once to produce the sequence-level representation.
On the second pass, we utilize cross attention and causal masking to learn the caption generation task. The text features can attend to the image features and predict the tokens in a sequence (see Figure~\ref{fig:two-pass}). As mentioned on the first pass we disable the cross-attention and causal masking in order to separate the text and image features for contrastive learning. All text-decoder weights are shared and gradients are aggregated from the two passes during training. The two passes are done interchangeably during training so their order is not important. 
As the two passes can share identical image representation, \ours achieves considerable computation savings compared to training two separate models.

We use a vision transformer as image encoder, and project the output dimension of image encoder to that of text decoder by a linear layer. The global image ($v_i$) and text representation ($l_i$) for contrastive learning are computed by average pooling over spatial dimensions and the sequence length, respectively. We insert $M$ cross-attention layers into $N$ text decoder layers, where $M \approx \frac{N}{2}$. The ratio $\frac{M}{N}$ represents a trade-off between model capacity and text-generative capability, where higher $M$ tends to benefit the text-generation tasks.

Somewhat surprisingly, \ours requires no task-specific predictor heads to unify the two seemingly disparate tasks other than a vocabulary embedding layer to map the decoder output features to text tokens. The same decoder features are average-pooled to represent the whole sequence ($l_i$). This is verified in our ablations as well.
Compared to contrastive captioner~\citep{yu2022coca}, \ours allows more flexible fusion of image and text features anywhere in the text decoder, and fully share the text-decoder parameters between contrastive and caption generation tasks.

\textbf{Model simplicity.}
In contrast to other architectures, Figure~\ref{fig:main_arch}, middle and right, we use a single vision encoder and a single text decoder. Unlike encoder-decoder models~\citep{pali,wang2022git,wang2022image} and contrastive alignment and captioning~\citep{albef,yu2022coca}, this is simpler and more flexible architecture. While using an encoder-decoder is popular, it comes with some disadvantages. For example, a pretrained visual encoder is  required, and contrastive tasks are usually not feasible. %, and adaptation to video is on a per-frame basis. % Additionally often an encoder-decoder is trained for text is adapted to an additional input.   
Contrastive alignment and captioning methods extend the encoder-decoder models to handle retrieval, but that mechanism is not easy to adapt to video and creates a hard loss-balancing task. 
\ours on the other hand works well for video-text tasks (Sec.~\ref{sec:video_sec}) and can be seamlessly adapted without extending the architecture, adding more encoders or having to use an encoder multiple times. 
%moved below
%In comparison, previous approaches some serving as Video Foundational Models, exhibit large complexity. InternVideo~\citep{internvideo} for example, supports a video masked encoder for MAE style losses in addition to an additional Video/Image Encoder and a Text decoder which are trained in the Align-Before-Fuse or Contrastive Captioner style, but for videos. Flamingo~\citep{flamingo} similarly is able to process either image and video inputs, but it trains a separate image-encoder, the text encoder/decoder part of which is discarded when integrating in the main model. We note that we outperform both in our experiments on video tasks.  

%\subsection{Natural Language Supervised Pretraining}
\subsection{Pretraining Losses}
Our model combines the strengths of popular vision-language pre-training losses, as are specifically described below, including with our modifications.

\paragraph{Contrastive Loss.}\quad Compared to the fully supervised classification pretraining, the contrastive learning approach uses separate image and text encoders to compute image and text embeddings of typically web-scale image-text pairs~\citep{radford2021clip,align,yu2022coca,zhai2021lit}. This allows to model to learn from the richer supervisory signal of free-form text than a fixed label set. The two encoders are trained jointly to minimize the contrastive objective~\citep{oord2018representation,radford2021clip,align}.
% \begin{equation}\label{eqn:i2t}
% L_{\text{I2T}} = -{1 \over {B}} \sum_{i=1}^{B} \log({\text{exp}(v_{i}l_{i} / \tau) \over { \sum_{ji=1}^{B} \text{exp}(v_{i} l_{j} / \tau)  }}). \quad L_{\text{T2I}} = -{1 \over {B}} \sum_{i=1}^{B} \log({\text{exp}(v_{i}l_{i} / \tau) \over { \sum_{j=1}^{B} \text{exp}(v_{ji} l_{ij} / \tau)  }}).
% \end{equation}
 As a result of learning from free-form text, the representation is very effective for zero-shot image classification, image-text retrieval, and robust to corrupted or out-of-distribution images~\citep{radford2021clip}.

\paragraph{Image Captioning loss.}\quad Dual-encoder contrastive models treat the text as a single entity to be encoded, while the encoder-decoder image captioner seeks a more detailed token-level understanding by predicting each word in sequence (see middle of Figure~\ref{fig:main_arch}). This is achieved through an encoder-decoder architecture, where the image encoder creates a latent representation of the image using ViT~\citep{dosovitskiy2020image} or ConvNet backbone. The text decoder then generates the tokens autoregressively by maximizing the likelihood of each predicted token given the previously generated tokens in the sequence, resulting in the following forward autoregressive formulation:
\begin{equation}\label{eqn:gen}
    L_{captioning}=-\sum_{t=1}^{T}\log P_{\theta}(y_t|y_{1, 2, ..., t-1}, x).
\end{equation}

To achieve maximum learning efficiency and parallelize computation, the encoder-decoder architecture is trained using a technique called teacher-forcing, which trains the model to predict the tokens at all time steps in parallel.

% Moving mostly to previous work and to the above architecture discussion. 

%\paragraph{Encoder-decoder Contrastive Captioner.}\quad Contrastive Captioner (CoCa)~\citep{yu2022coca} follows the standard image-text encoder-decoder architecture, where images are encoded into latent representations using a neural network encoder (such as ViT~\citep{dosovitskiy2020image}) and texts are encoded and decoded with a causal-masking transformer decoder. However, unlike typical decoder transformers, CoCa does not use cross-attention in the first half of the decoder layers. Instead, it encodes unimodal text representations and then use cross-attention to connect the text with the image representation to generate multimodal image-text representations, which is then fed through the remaining decoder layers  (see right of Figure~\ref{fig:main_arch}). This allows the decoder to produce both unimodal and multimodal text representations  by combining the contrastive and generative objectives as below:
%\begin{equation}\label{eqn:combine}
%    L_{total} = \lambda_{cap}L_{captioning} + \lambda_{con}L_{contrastive}.
%\end{equation}
%, where $\lambda_{cap}$ and $\lambda_{con}$ are loss balancing hyper-parameters.

\paragraph{Focal Contrastive Loss.}\quad
Contrastive learning typically relies on large batch size to extract supervisory signal from noisy image-text data. Our goal is to learn from the more informative and challenging examples than what is possible with the standard cross entropy loss. The focal loss~\citep{lin2017focal} presents a compelling alternative as it allows us to finely tune the weights assigned to challenging examples, demonstrating improved performances for object classification or detection scenarios. It has been recently shown, that applying focal loss achieves very competitive performance with significantly smaller batch size for contrastive learning~\citep{rovit}. More specifically the focal loss is applied as follows: Let $v_i$ and $l_i$ be the normalized image and text embeddings, and the image-to-text (I2T) focal contrastive loss be $L_\text{focal}$. We can write $L_\text{focal}$ mathematically as:
\begin{equation}\label{eq:focal}
L_{\text{focal\_contrastive}} = -{1 \over {B}} \sum_{i=1}^{B} \sum_{j=1}^{B} (1- p_{i})^{\gamma} \text{log}(p_{i}),
\end{equation}
where $p_i$ denotes the true class probability as below:
\begin{equation}
p_{i} =
    \begin{cases}
        \sigma(v_{i} l_{j} / \tau) & \text{if } i = j\\
        1 - \sigma(v_{i} l_{j} / \tau) & \text{if } i \neq j
    \end{cases}
\end{equation}
Here $\sigma$ denotes the sigmoid function, $v_i$ and $l_i$ the normalized image and text embeddings, and $\tau$ the learnable temperature to scale the logits. The loss is summed over the number of elements in the batch $B$.
where . For simplicity, we use the non-alpha-balanced focal loss~\citep{lin2017focal}. The total loss is the sum of I2T and T2I losses as follows:
\begin{equation}\label{eqn:contrastive}
L_{contrastive} = L_{\text{I2T}} + L_{\text{T2I}}.
\end{equation}
The focal contrastive loss modification provides additional sensitivity to objects, which equips the model for downstream detection tasks so we prefer it to the classical contrastive loss in our model.

\paragraph{Final loss} \quad Combining the losses above, our final loss is as follows. We note that while balancing of the losses is needed as is typical, the losses are simply combined and are shared with the full model.
\begin{equation}\label{eqn:combine}
    L_{total} = \lambda_{cap}L_{captioning} + \lambda_{focal}L_{\text{focal\_contrastive}},
\end{equation}
where $\lambda_{cap}$, $\lambda_{focal}$, are loss balancing hyper-parameters.

\paragraph{Learning from Scratch with Noisy Image-Text Supervision.}\quad Unlike many existing methods that train model components in multiple stages using different data sources or modalities~\citep{pali,flamingo,wang2022image,wang2022git}, \ours is pretrained end-to-end from scratch, without relying on any prior training or external sources. We use only a web alt-text dataset~\citep{align} for training. As the image encoder is typically the computation bottleneck in contrastive learning~\citep{radford2021clip}, our pretraining approach incurs only a relatively light overhead in training efficiency over a pure contrastive learner~\citep{radford2021clip} ($\approx 16\%$). This is highly desirable as scaling up model and data size have shown consistent benefits in contrastive and captioning pretraining~\citep{radford2021clip,pali,flamingo}.

%\paragraph{Learning Positional Embeddings for Open-Vocabulary Detection.}\quad
\paragraph{Learned Positional Embeddings for Localization Awarenesss.}\quad
Existing vision and language pretraining approaches and detection finetuning have a mismatch in how they use positional embeddings. Pretraining approaches typically use full-image positional embeddings during training and apply the same embeddings for downstream tasks. But for detection finetuning, recognition occurs at the region level, requiring the full-image positional embeddings to generalize to regions not seen during pretraining. 
To address this gap, we adopted the Cropped Positional Embedding~\citep{rovit}. The idea is to up-sample the positional embeddings from the pretraining image size (e.g., 224) to the detection task image size (e.g., 1024). Then, a randomly cropped and resized region from the up-sampled positional embeddings is used as the image-level positional embedding during pretraining. This method trains the model to view each image not as a full image, but as a region crop from a larger, unknown image, which better matches the downstream use case of detection where recognition occurs at the region level instead of the image level.

\begin{figure}
    \centering
     \vspace{-0.7cm}
    \includegraphics[width=0.37\linewidth]{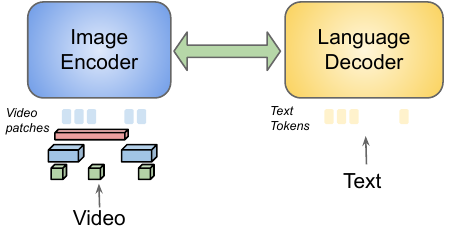}
    \hspace{0.5cm}
    \includegraphics[width=0.37\linewidth]{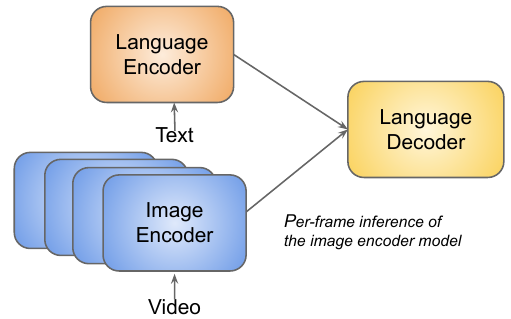}
     \vspace{-0.2cm}
    \caption{\ours video model (left) efficiently and seamlessly extends the image-language model, by adding learnable spatio-temporal features. The model is applied only once, as opposed to other models, processing each individual frame independently (right). Our model uses image-text pre-training only. }
    \label{fig:video_arch}
\end{figure}

\paragraph{Implementation Details.}\quad 
Our model consists of a standard ViT-Huge~\citep{dosovitskiy2020image,ScalingViTs} image encoder (of 650M parameters) and a transformer text decoder of 1B parameters. %(18 layers, model dimension 1536, hidden dimension 12288).   
The cross attention layers are applied every two decoder layers, where the model and hidden dimension follow the text decoder (420M params). Ablation results are conducted with ViT-Base image encoder and a smaller text decoder using 128M parameters, % (8 layers and model dimension 1024, hidden dimension 4096), 
and a smaller 4K batch size. The optimal hyper-parameters are then used to run the large model. The large model is trained for 500K steps using a batch size of 16K. We use AdamW optimizer with weight decay value 0.01. Our initial learning rate is 0.001, and both generative and contrastive loss weights are set to 1.0.  We first resize every image to 272x272 and randomly crop a 224x224 patch out for pretraining. We apply 10K warmup steps before applying linear LR decay to the end of training. The temperature in contrastive learning is learnable and initialized to 1.0. We use the standard Sentencepiece tokenizer and set the text length to 64 following existing works~\citep{align,yu2022coca}. We use an alt-text dataset of 1.8B image-text pairs~\citep{align}, as is common in prior methods. The dataset is used for both contrastive and generative pretraining. To better match the downstream detection task, we finetune the model for 100K iterations with the Cropped Positional Embedding for further localization awareness.

% Our model consists of a standard ViT-H (of 650M parameters) and a 1B text decoder with 8 (TBD) layers. Four cross attention weights are added across the model. The model is trained for TBD steps and TBD batch size. 
% Ablation results are conducted with a smaller version of ViT-Base and a 128M text decoder. The smaller-scale experiments runs for 500k steps and a 4K batch size.

% The model is jointly trained with the losses presented above with the following coefficients:.

% We train the model for extra 100K iterations with the Cropped Positional Embedding (CPE) for localization awareness that can better match the downstream detection task. 

% A small scale experiment is ran on the ViT-B level to determine the hyperparameters for the final pre-trained model and the fine-tuned versions.
% %Hyperparameters will be availble upon request.

%\subsection{\ours for Downstream Tasks}

\section{\ours for Video Tasks}
\label{sec:video_sec}
%Training video-text foundational models is a challenging task. One  approach is to leverage the image-language pre-trained model, where often the approach is to process individual frames through the encoder, which is not only slow, but also does not fully leverage the spatio-temporal information information from the video as higher order features are extracted from separate frames only. Many previous works also only use a small number of frames (e.g., 4 or 8).

Our video model is an efficient and seamless extension to the main image-language model, based on the TubeViT idea~\citep{piergiovanni2022tubevit}. It extracts video tubes which are then projected to patches similar to 2D image projections (Figure~\ref{fig:video_arch}, left). The model is applied only once. Some other image-language models adapted to video by processing each individual frame by the image encoder, e.g.~\citep{wang2022git,dynpretr,videococa} (Figure~\ref{fig:video_arch}, right). This is a limitation as only a relatively small number of frames can be processed due to memory and runtime constraints. There is also some evidence that for a large number of frames the number of effective tokens becomes too large which leads to deteriorated performance~\citep{piergiovanni2022tubevit}.

We follow the TubeViT approach~\citep{piergiovanni2022tubevit}. However, we apply several changes.
One main challenge is that TubeViT  requires fixed position embedding, whereas this does not match well with the learned positional embeddings of the main encoder. Since the video tubes are sparse and can overlap, TubeViT found that the fixed position embeddings were important. To enable those here, we propose using both position embeddings, and adding a weighted connection to the newly added fixed embeddings. Next, we use the same 2D patches, but at a sparse temporal stride, and finally add the tube projections following the settings used in TubeViT. All these tokens are concatenated together, and passed through the shared ViT backbone.
Our TubeViT adaptation to video is very lightweight and no additional components or losses are needed.
In comparison, previous approaches tend to exhibit higher complexity. InternVideo~\citep{internvideo} for example, supports a video masked encoder for MAE~\citep{He_2022_CVPR} losses in addition to a module similar to ALBEF. Flamingo~\citep{flamingo} is able to process either image and video inputs, but it trains a separate image-encoder, the text part of which is discarded when integrating in the main model. We note that we outperform these in our experiments on video tasks.
%In comparison to prior work, e.g.VindLU~\citep{VINDLU} the adaptation needs both changes in the main model, as well as thee additional video-text learning losses. 

We note that this change does not need any additional video data pre-training, and all experiments conducted in the paper are done by directly fine-tuning of the \ours image-text model on a video dataset. % We also compare adding further video training (e.g. VideoCC-3M \citep{}).

\vspace{-2mm}
\section{Experiments}
\vspace{-2mm}

%\begin{itemize}
%    \item Retrieval results
%    \item VQA results (high priority) + Post-processing.
%    \item Open-vocabulary detection results
%    \item Captioning results
%    \item Image classification results (optional)
%    \item Ablations on cross-attention and model capacity
%    \item Ablations on cross-task benefits.
%\end{itemize}

\subsection{Zero-Shot Image Retrieval Results}
\vspace{-0.2cm}

\label{sec:retrieval}

%\paragraph{Zero-shot Transfer}\quad TBD.
As Zero-shot image retrieval is a good indicator of the capability of the model without fine-tuning, we first evaluate the performance of \ours on Zero-shot image-text retrieval tasks.
Table~\ref{tab:retrieval_sota} shows the image-to-text and text-to-image results, compared to the  SOTA methods on two popular retrieval benchmarks MS COCO~\citep{chen2015cococaptions} and Flickr~\citep{plummer2015flickr30k}. As established by previous approaches, we evaluate both image-to-text and text-to-image retrieval following the same evaluation protocol and compare with other published dual-encoder models. Our approach significantly outperforms the state-of-the-art on both image-to-text and text-to-image retrieval by 2-4 points on the recall@1 metrics.
We note that these are challenging benchmarks many approaches have been evaluated on. % We omit earlier methods to save space. 

\begin{table*}[t]
\centering
\small
\tablestyle{5pt}{1.1}
\begin{tabular}{l|c|cccccc|cccccc}
% \toprule
&  image & 
\multicolumn{6}{c|}{MS COCO (5K test set)} & \multicolumn{6}{c}{Flickr30K (1K test set)} \\
& model & \multicolumn{3}{c}{\underline{{\white{-------}}image-to-text{\white{-------}}}} & \multicolumn{3}{c|}{\underline{{\white{-------}}text-to-image{\white{-------}}}} & \multicolumn{3}{c}{\underline{{\white{-------}}image-to-text{\white{-------}}}} & \multicolumn{3}{c}{\underline{{\white{-------}}text-to-image{\white{-------}}}} \\
% \hline
Method & size      & R@1  & R@5  & R@10 & R@1  & R@5   & R@10      & R@1  & R@5  & R10 & R@1  & R@5 & R@10    \\
\hline
CLIP~\citep{radford2021clip}   & 302M
& 58.4 & 81.5 & 88.1 & 37.8 & 62.4 & 72.2       & 88.0 & 98.7 & 99.4 & 68.7 & 90.6 & 95.2        \\
ALIGN~\citep{align}   & 408M
& 58.6 & 83.0 & 89.7 & 45.6 & 69.8 & 78.6       & 88.6 & 98.7 & 99.7 & 75.7 & 93.8 & 96.8        \\
FLAVA~\citep{singh2022flava}   & 86M
& 42.7 & 76.8 & - & 38.4 & 67.5 & -             & 67.7 & 94.0 & - & 65.2 & 89.4 & - \\
FILIP~\citep{yao2021filip}   & 302M
& 61.3 & 84.3 & 90.4 & 45.9 & 70.6 & 79.3       & 89.8 & 99.2 & 99.8 & 75.0 & 93.4 & 96.3        \\
Florence~\citep{yuan2021florence} & 637M
& 64.7 & 85.9 & - & 47.2&  71.4&  -             & 90.9 & 99.1 & - & 76.7 & 93.6 & -             \\
% \rowfont{\color{lightgray}}
% Flamingo& 65.9 & 87.3 & 92.9 & 48.0 & 73.3 & 82.1        & 89.3 & 98.8 & 99.7 & 79.5 & 95.3 & \bf{97.9}       \\
% \rowfont{\color{lightgray}}
CoCa-L~\citep{yu2022coca} & 303M
& 65.4 & 85.6 & 91.4&  50.1 & 73.8 & 81.8          & 91.4 & 99.2 & \bf{99.9} & 79.0& 95.1 & 97.4   \\
CoCa~\citep{yu2022coca} & 1B
& 66.3 & 86.2 & 91.8 & 51.2 & 74.2 & 82.0          & 92.5 & 99.5 & \bf{99.9} & 80.4 & 95.7 & 97.7 \\
\hline
\bf{\ours (ours)}   & 630M  & \bf{70.7}  & \bf{89.1}  & \bf{93.7}  & \bf{54.1}  & \bf{76.8}  & \bf{84.2}
        & \bf{94.9}  & \bf{99.5}  & \bf{99.9}   & \bf{82.5}  & \bf{96.0}  & \bf{98.0}  \\
\hline
% \bottomrule
\end{tabular}
\vspace{-1mm}
\caption{\textbf{Zero-shot image-text retrieval results on COCO and Flickr30K benchmarks (dual-encoder models).} We evaluate our pretrained model compared to other methods. We achieve state-of-the-art results by a margin the on image-to-text / text-to-image retrieval benchmarks with comparable model capacity. (\textbf{bold}: best).}
\label{tab:retrieval_sota}
\vspace{-2mm}
\end{table*}

\vspace{-0.3cm}
\subsection{Visual Question Answering Results}
\vspace{-0.2cm}
\label{sec:vqa}

We report the performance on the VQAv2 benchmark~\citep{agrawal2015vqa} in Table~\ref{tab:results_vqa}.
Inspired by recent VQA approaches~\citep{pali,flamingo,piergiovanni2022answer}, we conduct the experiments in the open-ended text generative setting using an English vocabulary size of 256K. Most prior approaches~\citep{wang2021simvlm,yu2022coca,wang2022image,wang2022unifying} address the VQA task in the classification setting where the best answer is selected from a predefined set of answers (typically of size 3K). Some recent works~\citep{albef,li2022blip} train the model in an open-ended settings but restrict the decoder to generate only the 3K candidate answers during inference. In contrast, we allow the decoder to use the whole vocabulary during inference. The VQA-as-open-generation setting poses two key challenges: firstly, the produced text must precisely match the expected answer in order to be considered correct, and secondly, our vocabulary size is much larger than the ones utilized in the classification settings.

\ours achieves 80.8 accuracy on this benchmark, which is very competitive among the open-ended generation approaches. For example, \ours outperforms the PaLI-3B by 1.4 points, while using 1.5x fewer parameters (2B total params). Compared to the PaLI-15B, ours achieves the same performance while being 7.5x smaller. In addition, Flamingo and PaLI use a combination of interleaved image-text, alt-text, human-annotated data sources for training~\citep{flamingo,pali}, whereas we use only an alt-text dataset~\citep{align}. Larger models, such as PaLI-17B, outperform ours. 
Table~\ref{tab:results_vqa_analysis} shows \ours performance on VQA test-std set by question types. \ours performs the best on yes/no question type, and less well on questions that require counting.
%Finally, \ours achieves similar accuracy to strong closed-vocabulary approaches such as SimVLM~\citep{wang2021simvlm} and Florence~\citep{yuan2021florence}.

\begin{table}[t]
\centering
\small
%\tablestyle{6pt}{1.1}
\begin{tabular}{l|c|c}
% \toprule
% Method   & Overall &Yes/No & Number & Other \\ 
Method & Test-Dev & Test-Std \\ 
\hline
FLAVA~\citep{singh2022flava} & \gray{72.8} & - \\
METER~\citep{meter} & \gray{77.7} & \gray{77.6} \\
Unified-IO~\citep{UnifiedIO} & \gray{77.9} &- \\
OmniVL~\citep{omniVL} & \gray{78.3} & \gray{78.4} \\
Florence~\citep{yuan2021florence} & \gray{80.2} & \gray{80.4} \\
SimVLM~\citep{yu2022coca} & \gray{80.0} & \gray{80.3} \\
OFA~\citep{wang2022unifying} & \gray{82.0} & \gray{82.0} \\
CoCa~\citep{yu2022coca} & \gray{82.3} & \gray{82.3} \\
BEiT-3~\citep{yu2022coca} & \gray{84.2} & \gray{84.0} \\
\hline
% Open-ended generation: \\
ALBEF~\citep{albef}  & 75.8 & 76.0 \\
AnswerMe~\citep{piergiovanni2022answer} & 73.6 & - \\
BLIP~\citep{li2022blip}  & 78.2 & 78.3 \\      
GIT~\citep{wang2022git}  & 78.6 & 78.8 \\
%Dynamic Pretr.~\citep{dynpretr} & 79.9 & - \\
Flamingo-80B~\citep{flamingo} & 82.0 &  82.1 \\
BLIP-2-7B~\citep{Li2023BLIP2BL} & 82.3 &  - \\
PaLI-3B~\citep{pali}   & 79.3 & - \\
PaLI-15B~\citep{pali}   & 80.8 & - \\
PaLI-17B~\citep{pali}   & 84.3 & 84.3 \\
\hline
\bf{\ours (2B)} & 80.7 & 80.8 \\
\hline
\end{tabular}
\caption{\textbf{Visual Question-Answering (VQAv2).} We benchmark the performance in an open-ended generation setting. Approaches that perform VQA in closed-vocabulary settings are marked in \gray{gray}. \ours is very competitive among existing open-ended generation methods given its modest model capacity.
}
\label{tab:results_vqa}
\end{table}

\begin{table}%[t]
\centering
\small
\begin{tabular}{c|ccc}
Overall &Yes/No & Number & Other \\ 
\hline
80.84 & 93.41 & 63.89 & 73.78 \\
\hline
\end{tabular}
\vspace{-1mm}
\caption{\textbf{Visual Question-Answering (VQAv2) analysis.} Performance by question types on the test-std split. %"Yes/No" questions, not surprisingly, perform best, whereas "Number" questions, answering for how many objects are present, are the most challenging.
"Yes/No" questions perform best, whereas questions about how many objects are present, are the most challenging.
}
\vspace{-5mm}
\label{tab:results_vqa_analysis}
\end{table}

 \begin{comment}
 
 \subsection{Image Captioning Results}
 \label{sec:cap}

Table~\ref{tab:results_cap} shows \ours performance on the Image Captioning task on the MS-COCO Captioning benchmark~\citep{chen2015cococaptions}. 
Similarly, to the results on VQA, here too, models such as PaLI~\citep{pali} and GIT/GIT2~\citep{wang2022git} outperform our method.

 \begin{table}[t]
 \centering
 \small
 %\tablestyle{6pt}{1.1}
 \begin{tabular}{l|c|c}
 % \toprule
 Method   & CIDEr &Bleu \\ 
 \hline
 BLIP~\citep{li2022blip} &136.7 &40.4\\
 Flamingo-80B~\citep{flamingo} &138.1 &- \\
 %LEMON-Base~\citep{lemon} &133.3 &40.3\\
 LEMON-Huge~\citep{lemon} &139.1 &41.5\\
 VinVL~\citep{VinVL} &140.9 & 41.0\\
 CoCa~\citep{yu2022coca} &143.6 &40.9\\
 GIT~\citep{wang2022git} &144.8 &44.1 \\
 GIT2~\citep{wang2022git} &145.0 &\textbf{44.1} \\
 %OFA~\citep{wang2022unifying} &145.3 & \\
 PaLI-3B~\citep{pali} &145.4 &- \\
 PaLI-17B~\citep{pali} &\textbf{149.1} &- \\
 \hline
 \bf{\ours (ours)}   & 136.8 & 38.3 \\
 \hline
 % \bottomrule
 \end{tabular}
 \caption{\textbf{Image Captioning.} Performance on the Image Captioning Dataset evaluated on the MS-COCO dataset.}
 \label{tab:results_cap}
 \vspace{-2mm}
 \end{table}

 \end{comment} %% comment

\vspace{-1mm}
\subsection{Video Question Answering Results}
\label{sec:video}

In this section we present the results of our model on the Video Question Answering task, which is a challenging task answering questions about activities, events, objects, or repetition counting within a video. Our VideoQA results are obtained using the image-text pre-trained model, namely we directly fine-tune on a VideoQA dataset without any video-text pre-training. This is in contrast with other works which use video data pre-training, and indicates that our model is already very strong without additional video-text pre-training.

The results are presented in Table~\ref{tab:video_qa}, comparing to the state of the art on MSRVTT-QA~\citep{xu2016msrvtt} and MSVD-QA~\citep{xu2017msvd-qa} datasets. \ours outperforms the best SOTA approaches, among which are both video foundational models, e.g VIOLET, MERLOT, InternVideo, image-text models adapted to video e.g. GIT and GIT2~\citep{wang2022git}, and large vision-language models, such as Flamingo~\citep{flamingo}.
We note that, similar to the VQA results, our VideoQA results are conducted in the more challenging open-ended generation setting.
Our model is 2.5 times smaller than GIT2 (5B parameters), and 40 times smaller than Flamingo (80B parameters).

% \begin{table}[t]
% \centering
% \small
% \tablestyle{6pt}{1.1}
% \begin{tabular}{l|c|c|c}
% Method	&MSRVTT-QA &MSVD-QA & \\
% \hline

% Just Ask~\citep{yang2021justask}	&41.5 &46.3 & \\
% MERLOT~\citep{merlot}	&43.1 &- & \\
% OmniVL~\citep{omniVL} &44.1 &51.0 & \\
% VindLU~\citep{VINDLU} & 44.6 & - \\
% Iterative Co-Tok~\citep{piergiovanni2022cotok}	&45.7 &48.8 & \\
% All-in-one~\citep{wang2022allinone}  &46.8 &48.3 & \\
% Video-Coca~\citep{videococa}	&46.3 &56.9 & \\
% VIOLET~\citep{fu2021violet} &43.9 &47.9 & \\
% VIOLETv2~\citep{fu2023empiricalmvm} &44.5 &54.7 & \\
% Dynamic Pretr.~\citep{dynpretr} & 45.1 & 47.1 \\
% GIT~\citep{wang2022git} &43.2 &56.8 & \\
% GIT2~\citep{wang2022git} &45.6 &58.2 & \\
% %PaLI-3 	&47.1 &- & \\
% InternVideo~\citep{internvideo} &47.1 &55.5 & \\
% Flamingo~\citep{flamingo}	&47.4 &- & \\
% %\shline
% %M-PLUG2~\citep{mPLUG-2}	&48.0  &58.1 & \\
% \hline
% %\textbf{\ours (ours)} 	&\textbf{49.5} &\textbf{59.36} & \\ 
% \textbf{\ours (ours)} 	&\textbf{49.5} &\textbf{60.2} & \\ 
% \hline
% \end{tabular}
% \vspace{2mm}
% \caption{\textbf{Video QA Results.} We report performance on Video Question Answering. Our approach outperforms the SOTA on both MSRVTT-QA and MSVD-QA datasets. We note that image-language pre-training is the only one used here. \ours outperforms both video-first models e.g. VIOLET, InternVideo, image+video models, e.g. Flamingo, as well as, large pre-trained image-language models adapted to video, e.g. GIT, GIT2.}
% \label{tab:video_qa}
% \vspace{-2mm}
% \end{table}

\subsection{Video Captioning Results}
\label{sec:video_cap}
Video captioning results on the MSRVTT~\citep{xu2016msrvtt} and MSVD~\citep{msvd,xu2017msvd-qa} datasets are presented in Table~\ref{tab:video_cap}. 
 The results here too are obtained with image-text pre-training only. Our approach performs well related to SOTA. It outperfoms prior approaches on the MSVD benchmark by large margins, and outperforms most others with the exception of GIT/GIT2 on the MSRVTT video captioning dataset.

% \begin{table}[t]
% \centering
% \small
% \tablestyle{6pt}{1.1}
% \begin{tabular}{l|c|c}
% Method	&MSRVTT   & MSVD \\
% \hline
% %UNI-VL~\citep{UniVL} &50.0 &- \\
% ORG-TRL~\citep{orgtrl} &50.9 &95.2  \\
% SWINBert~\citep{swinbert} &53.8 &120.6 \\
% VIOLETv2~\citep{fu2023empiricalmvm} &58.0 &130.2  \\
% MV-GPT~\citep{mvgpt} &60.0 &- \\
% Vid2Seq~\citep{vid2seq} &64.6 & 146.2\\
% Video-Coca~\citep{videococa}	&73.2 & - \\
% %PaLI-3 	&76.8 \\
% GIT~\citep{wang2022git} &73.9 &180.2 \\
% GIT2~\citep{wang2022git} &\textbf{75.9} &185.2 \\
% %\shline
% %M-PLUG2~\citep{mPLUG-2}	&48.0  &58.1 & \\
% \hline
% \textbf{\ours (ours)} 	&73.6 &\textbf{195.6} \\ 
% \hline
% \end{tabular}
% \vspace{2mm}
% \caption{\textbf{Video Captioning Results.} %Results on the Video Captioning task. 
% \ours performs well on both MSRVTT and MSVD Video Captioning Benchmarks, outperforming SOTA on MSVD by large margins. CIDEr scores are shown.}
% \label{tab:video_cap}
% \vspace{-2mm}
% \end{table}

\begin{table}[h]
    \begin{subtable}[h]{0.56\textwidth}
        \centering
        \small
        \tablestyle{6pt}{1.1}
        \begin{tabular}{l|c|c}
        Method	&MSRVTT-QA &MSVD-QA \\
        \hline
        
        Just Ask~\citep{yang2021justask}	&41.5 &46.3 \\
        MERLOT~\citep{merlot}	&43.1 &- \\
        OmniVL~\citep{omniVL} &44.1 &51.0 \\
        VindLU~\citep{VINDLU} & 44.6 & - \\
        Iterative Co-Tok~\citep{piergiovanni2022cotok}	&45.7 &48.8 \\
        All-in-one~\citep{wang2022allinone}  &46.8 &48.3 \\
        Video-Coca~\citep{videococa}	&46.3 &56.9 \\
        VIOLET~\citep{fu2021violet} &43.9 &47.9  \\
        VIOLETv2~\citep{fu2023empiricalmvm} &44.5 &54.7  \\
        Dynamic Pretr.~\citep{dynpretr} & 45.1 & 47.1 \\
        GIT~\citep{wang2022git} &43.2 &56.8  \\
        GIT2~\citep{wang2022git} &45.6 &58.2  \\
        %PaLI-3 	&47.1 &- & \\
        InternVideo~\citep{internvideo} &47.1 &55.5 \\
        Flamingo~\citep{flamingo}	&47.4 &- \\
        %\shline
        %M-PLUG2~\citep{mPLUG-2}	&48.0  &58.1 & \\
        \hline
        %\textbf{\ours (ours)} 	&\textbf{49.5} &\textbf{59.36} & \\ 
        \textbf{\ours (ours)} 	&\textbf{49.5} &\textbf{60.2} \\ 
        \hline
        \end{tabular}
        \vspace{2mm}
        \caption{\textbf{Video QA Results.} \ours outperforms the SOTA on both MSRVTT-QA and MSVD-QA datasets. We note that image-language pre-training is the only one used here. \ours outperforms video-first models e.g. VIOLET, InternVideo, image+video models, e.g. Flamingo, and large pre-trained image-text models adapted to video, e.g. GIT2.}
        \label{tab:video_qa}
    \end{subtable}
    \hfill
    \begin{subtable}[h]{0.4\textwidth}
        \centering
        \small
        \tablestyle{6pt}{1.1}
        \begin{tabular}{l|c|c}
        Method	&MSRVTT   & MSVD \\
        \hline
        %UNI-VL~\citep{UniVL} &50.0 &- \\
        ORG-TRL~\citep{orgtrl} &50.9 &95.2  \\
        OpenBook~\citep{openbook} &52.9 &- \\
        SWINBert~\citep{swinbert} &53.8 &120.6 \\
        VIOLETv2~\citep{fu2023empiricalmvm} &58.0 &130.2  \\
        MV-GPT~\citep{mvgpt} &60.0 &- \\
        Vid2Seq~\citep{vid2seq} &64.6 & 146.2\\
        Video-Coca~\citep{videococa}	&73.2 & - \\
        %PaLI-3 	&76.8 \\
        GIT~\citep{wang2022git} &73.9 &180.2 \\
        GIT2~\citep{wang2022git} &\textbf{75.9} &185.2 \\
        %\shline
        %M-PLUG2~\citep{mPLUG-2}	&48.0  &58.1 & \\
        \hline
        \textbf{\ours (ours)} 	&73.6 &\textbf{195.6} \\ 
        \hline
        \end{tabular}
        \vspace{2mm}
        \caption{\textbf{Video Captioning Results.} %Results on the Video Captioning task. 
        \ours performs well on both MSRVTT and MSVD Video Captioning Benchmarks, outperforming SOTA on MSVD by large margins. CIDEr scores are shown. As for VideoQA experiments, we use only image-language pre-training, and directly fine-tune the model on each dataset.}
        \label{tab:video_cap}
     \end{subtable}
     \caption{Video Question Answering (VideoQA) and Video Captioning results.}
     \vspace{-3mm}
     \label{tab:temps}
\end{table}

\subsection{Open-vocabulary Detection Results}
\label{sec:detection}

The Open-Vocabulary detection task refers to the ability to detect and name objects (providing bounding boxes) of categories that are unknown to the model. We evaluate our work on the challenging LVIS dataset~\citep{lvis} which features 1200+ different object categories. We report performances using the Average Precision (AP) metrics for rare classes, as previously established in the literature~\citep{gu2022openvocabulary,minderer2022simple}. 
The open-vocabulary detector is initialized with the pretrained ViT backbone during finetuning. %, as in~\citep{kuo2022fvlm,rovit}. 
It adopts the simple feature pyramid and windowed attention to handle higher resolution images (e.g., 1024) as proposed in ViTDet~\citep{li2022exploring}, and Mask R-CNN heads and class-agnostic box/mask heads as in~\citep{du2022learning,gu2022openvocabulary,Zareian_2021_CVPR,zhong2021regionclip}. The model is trained with base categories, and tested to detect novel category objects at inference.

Table~\ref{tab:ovd} presents our results on Open-Vocabulary detection which evaluates how well it does on detecting novel, previously unseen object categories. \ours achieves 31.0 AP$_r$, which outperforms the existing ViT-based method OWL-ViT by 5.4 points. %, and is very competitive to the SOTA approach F-VLM.
We also report AP performance for all classes for context, but note that our model is not optimized for detection AP, as is for example OWL-ViT which is a detection-only model.

\begin{table}[t]
\centering
\small
\tablestyle{8pt}{1.1}
\begin{tabular}{l|c|c}
% \toprule
%& \multicolumn{2}{c|}{Flickr30K} & \multicolumn{2}{c|}{MS COCO} & \multicolumn{1}{c}{VQA} \\
%\# Cross-Att. & I2T & T2I & I2T & T2I & Acc. \\ 
Method	&APr &AP  \\
\hline
DetPro-Cascade~\citep{du2022learning}  & 20.0 &\gray{27.0}\\
Detic-CN2~\citep{zhou2022detecting} & 24.6 &\gray{32.4}\\
RegionCLIP~\citep{zhong2021regionclip} & 22.0 &\gray{32.3}\\
%ViLD-Ens & 18.7 &26.0\\
ViLD-Ensemble~\citep{gu2022openvocabulary} & 21.7 &\gray{29.6}\\
ViLD-Ensemble~\citep{gu2022openvocabulary} & 26.3 &\gray{29.3}\\
VL-PLM~\citep{zhao2022exploiting} & 17.2 &\gray{27.0}\\
Rasheed et al.~\citep{rasheed2022bridging} & 21.1 &\gray{25.9} \\
OWL-ViT~\citep{minderer2022simple} & 23.3 & \textbf{\gray{35.3}}\\
OWL-ViT~\citep{minderer2022simple} &25.6 &\gray{34.7}\\
% since the paper is not out the results are not technically official so we shouldn't "announce" them here so it is best to not report it in the table, we cite it as method thou in the paper which is fine. 
%RO-ViT~\citep{rovit} & 32.1 & 34.0 \\
%F-VLM~\citep{kuo2022fvlm} & 32.8 &34.9 \\
\hline
%\textbf{\ours (ours)-650M} 	& 30.2 & 34.6 \\ 
\textbf{\ours (ours)} 	& \textbf{31.0} & \gray{32.8} \\ 
%\hline
%\textbf{\ours (ours)} &31.0 & 33.0 \\
\hline
\end{tabular}
\vspace{2mm}
\caption{\textbf{Open-Vocabulary Detection Results.} \ours performs well on detecting novel objects, scoring much higher on the Average Precision (AP) of rare objects $AP_r$. 
%, which is standard Open-Vocabulary detection). %We report the Average Precision (AP) of rare objects $AP_r$, which measures the Open-Vocabulary performance, and the overall AP.
}
\label{tab:ovd}
\vspace{-2mm}
\end{table}

%------------------------------------------------------------------------

\section{Ablation studies}
\label{sec:ablation}

We here present ablation studies  to understand the model characteristics and its design choices. We use a ViT-Base image encoder and a 128M language decoder, training on 4K batch size for 500K iterations unless noted otherwise.

%\subsection{Model design ablations}

\textbf{Cross-task benefits.}
We first explore cross-task benefits between contrastive and text-generative pretraining. We find that joint training  is generally favorable to tasks, but it affects tasks differently (Table~\ref{tab:ablation_cross_tasks}). Compared to the contrastive-only pretraining, joint training achieves signficantly better VQA performance by +8 points, which is expected. Joint training improves text-to-image retrieval. This is likely because the generative modeling enhances text representation, and while remaining very competitive on the image-to-text retrieval, it is not as beneficial, which can indicate potential competition of tasks (please see our later experiments which explore this). Compared to the generative-only pretraining, joint training outperforms on VQA by +1.8 points, likely because the contrastive learning helps to improve the joint image-text representation. We attribute this to better representation learning rendered from contrastrive training, also evidenced by other works~\citep{radford2021clip,align}. More importantly, joint training model is able to tackle retrieval tasks, not supported by the generative pretraining, and achieving satisfying performance overall.

\begin{table}%[t]
\centering
\small
\tablestyle{8pt}{1.1}
\begin{tabular}{c|c|cc|cc|c}
% \toprule
&   & 
\multicolumn{2}{c|}{MS COCO} & \multicolumn{2}{c|}{Flickr30K} & \multicolumn{1}{c}{VQA} \\
Contrastive & Generative & I2T & T2I &I2T &T2I &Acc. \\ 
\hline
%\checkmark  &   & \bf{82.6}    & 67.1  & \bf{54.8}    & 38.2    & 63.5 \\
\checkmark  &   & \bf{54.8}    & 38.2  & \bf{82.6}    & 67.1    & 63.5 \\

 & \checkmark   & - & - & - & - & 69.9 \\
%\checkmark  & \checkmark   &  80.6 & \bf{67.5} & 54.3 & \bf{38.7} &
\checkmark  & \checkmark   &  54.3 & \bf{38.7} & 80.6 & \bf{67.5} &
71.7 \\
\hline
\end{tabular}
\vspace{2mm}
\caption{\textbf{Cross-task benefits.} Combining the contrastive and generative objectives yield benefits for generative task in our setup, while maintaining the performance of discriminative tasks.}
\label{tab:ablation_cross_tasks}
\vspace{-2mm}
\end{table}

\textbf{Cross-attention Design.}
%\label{sec:cross}
Another key exploration is the role of cross-attention mechanisms in the proposed joint training and how they affect various tasks. Cross-attention provide an efficient means for communication between the two modalities.
We find that tasks indeed perform better under different circumstances. % and a single model is hard to reconcile for these tasks. 
Specifically,  cross-attention is preferred for text generative tasks, but not as much for contrastive ones. Table~\ref{tab:ablation_cross_att} shows that while denser cross-attention connections (e.g. 4) benefit the VQA task, one or few layers (e.g. 2) are sufficient for retrieval task. 

\begin{table}%[t]
\centering
\small
\tablestyle{8pt}{1.1}
\begin{tabular}{c|cc|cc|c}
% \toprule
& \multicolumn{2}{c|}{MS COCO} & \multicolumn{2}{c|}{Flickr30K} & \multicolumn{1}{c}{VQA} \\
\# Cross-Att. & I2T & T2I & I2T & T2I & Acc. \\ 
\hline
%1 & 81.7 & 67.2 & 55.3 & 39.6 & 68.7 \\
1 & 55.3 & 39.6 & 81.7 & 67.2  & 68.7 \\
%2 & 81.9 & \bf{67.6} & \bf{56.6} & \bf{40.1} & 70.8 \\
2 &  \bf{56.6} & \bf{40.1} &81.9 & \bf{67.6}  & 70.8 \\
%4 & \bf{82.2} & 67.3 & 55.7 & 39.9 & \bf{71.5} \\
4 & 55.7 & 39.9 & \bf{82.2} & 67.3  & \bf{71.5} \\
\hline
\end{tabular}
\vspace{-1mm}
\caption{\textbf{Cross-attention Design.} Denser cross-attention layers are beneficial for the VQA task, whereas few cross-attention layers are sufficient for retrieval tasks.}
\label{tab:ablation_cross_att}
\vspace{-2mm}
\end{table}

\textbf{Balancing contrastive-vs-generative losses.}
%\label{sec:weights}
We here explore how to train the joint objectives, so that the tasks benefit fully from joint training. We do observe a potential competition of these tasks, thus a balance between losses needs to be obtained.
Table~\ref{tab:ablation_balancing} shows the performance trade-off when training jointly. %Here we explore balancing the contrastive vs generative losses to isolate other effects. 
The experiment is done on a smaller model and fewer steps, as mentioned above.
We observe that the two objectives indeed have competitive behaviors. In the rest of the experiments, we pick equal weights over these two losses, as we observe that it gives more advantage to the VQA performance, whereas the retrieval does not suffer as much. This experiment provides an insight as to how to tune these parameters, depending on the requirements of the application.

\begin{table}%[t]
\centering
\small
\tablestyle{8pt}{1.1}
\begin{tabular}{c|cc|cc|c}
% \toprule
& 
\multicolumn{2}{c|}{MS COCO} & \multicolumn{2}{c}{Flickr30K} & \multicolumn{1}{c}{VQA} \\
 weights & I2T & T2I &I2T &T2I &Acc. \\ 
\hline
(2.0, 0.5)  & \bf{56.71}    & \bf{40.77}  
        & \bf{82.52}    & \bf{67.77}    & 70.08\\
(2.0, 1.0)  & 56.7   & 40.27  
        & 81.84   & 67.25  & 70.84 \\
(1.0, 1.0)  & 56.25   & 39.73  
        & 81.74   & 67.50  & 71.48 \\
(1.0, 2.0)  & 55.39   & 39.09  
        & 81.54   & 65.64  & \bf{72.27} \\
(0.5, 2.0)  & 52.05   & 37.32  
        & 78.32  & 62.73  & 71.79 \\
\hline
% \bottomrule
\end{tabular}
\vspace{-1mm}
\caption{\textbf{Balancing the losses.} Zero-shot retrieval results on COCO and Flickr30K benchmarks (R@1 shown) and VQA accuracy. Weight coefficients for the contrastive and generative loss are denoted as (constrastive, generative). We observe a clear trade-off between retrieval vs VQA tasks. We chose (1.0, 1.0) to balance the two objectives.
}
\label{tab:ablation_balancing}
\vspace{-1mm}
\end{table}

\begin{table}%[t]
\centering
\small
\tablestyle{8pt}{1.1}
\begin{tabular}{c|cc|cc}
% \toprule
& \multicolumn{2}{c|}{MS COCO} & \multicolumn{2}{c}{Flickr30K} \\
Image Tower Size & I2T & T2I &I2T &T2I \\ 
%\multicolumn{3}{c}{\underline{{\white{-------}}image-to-text{\white{-------}}}} & \multicolumn{3}{c|}{\underline{{\white{-------}}text-to-image{\white{-------}}}} & \multicolumn{3}{c}{\underline{{\white{-------}}image-to-text{\white{-------}}}} & \multicolumn{3}{c}{\underline{{\white{-------}}text-to-image{\white{-------}}}} \\
% \hline
%method & size      & R@1   & R@1       & R@1  & R@1      \\
\hline
% CLIP~\citep{radford2021clip}   & 302M
% & 58.4  & 37.8        & 88.0  & 68.7         \\
% ALIGN~\citep{align}   & 408M
% & 58.6  & 45.6        & 88.6  & 75.7         \\
% CoCa~\citep{yu2022coca} & 1B
% & 66.3  & 51.2           & 92.5  & 80.4 \\
% \hline
86M  & 62.0  & 44.2  & 84.6   & 71.1  \\
300M  & 66.4  & 49.4  & 91.2    & 78.8   \\
630M  & \bf{70.7}   & \bf{54.1} & \bf{94.9}   & \bf{82.5}  \\
\hline
% \bottomrule
\end{tabular}
\vspace{-1mm}
\caption{\textbf{Model scaling.} We show clear improvements on image-text retrieval with increasing image tower capacity. }
\label{tab:ablation_scale}
\vspace{-2mm}
\end{table}

\textbf{Scaling Image Encoder.}
Effects of model scale on vision-language tasks have been explored in prior works~\citep{flamingo,pali}. We present the results of scaling image encoder in Table~\ref{tab:ablation_scale}, where we confirm increasing the capacity of image encoder yields consistent improvement. We note that text encoder size of the last row has 1B parameters as opposed to 128M in the rest of the table. We use 16K batch size in this ablation.   %Furthermore, we observe that even small models very competitive with larger SOTA models. This is due to the benefits our our approach (for direct comparison, we have used the same dataset~\citep{align} as SOTA methods~\citep{yu2022coca}.). 

%\subsubsection{Tube-ViT Adaption Experiments}
%\subsubsection{Video Tubes Adaption Experiments}

\begin{table}
\centering
\small
\tablestyle{8pt}{1.1}
    \begin{tabular}{l|cc}
         & MSRVTT-QA & MSVD-QA \\
         \hline
        \ours - Full Model & \textbf{42.1} & \textbf{45.8} \\
        No Gated Connection & 41.8 & 45.5  \\
        No Fixed Embeddings & 41.5 & 45.1  \\
        No Tubes & 40.3 & 42.6 \\
  \hline
    \end{tabular}
    \vspace{-1mm}
    \caption{\textbf{Video Tubes Adaption Experiments.} Comparing different effects of adapting the model to video. }
    \label{tab:video_ablation}
    \vspace{-1mm}
\end{table}

\textbf{Video Model Ablations.}
In Table \ref{tab:video_ablation}, we compare the different design choices for adaptation of the video model. We explore the effects of removing gated connections, the fixed embeddings and even the video feature inputs. As seen, their importance increases in that order, where the addition of video tubes to process the video are of highest importance. This is not surprising as they learn the spatio-temporal video information. These ablations are done with the smaller ViT-L model. %The ablations here are done with a slightly smaller model, more specifically, we use the ViT-L, instead of ViT-H. 

\vspace{-2mm}
\section{Broader impacts}
\vspace{-1mm}
Developing vision-language models is very beneficial as they have many potential uses and applications, being able to understand both visual and language inputs better and to more accurately respond to  questions. %Such models can have further applications in Robotics and other related fields.
Some aspects of the model, specifically the text generative components might exemplify certain risks of generating off-topic, stereotypical, unwanted or other types of outputs, for which further investigation is needed.
Our model has been trained on the same image-text dataset as many other previous works ~\citep{align,ScalingViTs,zhai2021lit,yu2022coca,videococa} and is collected as noisy image-language pairs in a similar fashion to an even broader set of methods in the literature~\citep{pali,yuan2021florence,flamingo}. 
%Our model further demonstrates excellent performance on video-language tasks, while still being pre-trained on a single image-text dataset. This provides a further advantage of leveraging a single data source. 
We developed the model for exploring novel research capabilities and have only used it for evaluation and visualization purposes in this paper. 
\vspace{-2mm}
\section{Conclusions}
\vspace{-1mm}
We present the \ours model which is a vision-encoder text-decoder model capable of multiple vision-language tasks. 
We propose two-pass learning which allows us to train jointly for retrieval and text-generative tasks with fully shared weights. %Zero-shot retrieval results outperform the state-of-the-art (SOTA) models. Fine-tuning on several tasks including open-vocabulary object detection shows competitive of above SOTA results 
The model is easy to adapt to video-language and object detection tasks.
Our model accommodates a set of diverse tasks, such as image-text and text-image retrieval, novel (or open-vocabulary) object detection, VQA, VideoQA and Video Captioning, obtaining very strong performance on them with respect to the state-of-the art.
%The model is also easy to adapt to video-language tasks, such as Video Question Answering, outperforming the SOTA results, many of which are video-pretrained Foundational models, despite its smaller size and image-text only pre-training. %While not all categories of tasks are evaluated the model is capable to perform other tasks such as Image and Video Reasoning, Video Captioning, etc.  

%We do not intend to use it for other purposes, before further 

%\section*{Acknowledgements}
%We would like to thank Mojtaba Seyedhosseini, Vijay Vasudevan,  Priya Goyal, Qingqing Huang, Andy Ly, Jiahui Yu, Zirui Wang, Yonghui Wu, Runze Li, Jie Mei, Nan Du, Yuxin Wu, Radu Soricut, Tom Duerig, Paul Natsev, Zoubin Ghahramani for their help and support.

\bibliography{egbib}
\bibliographystyle{tmlr}

\appendix
\section{Appendix}
%You may include other additional sections here.

\subsection{Additional ablations}

We here include additional ablation experiments which can provide further introspection into the model.

\textbf{Language pre-training effects.}
As shown in this paper, a purely language model can be easily adapted for multiple image-language tasks. We here explore the effects of language pre-training on the vision-language tasks. Our results show that language-only pre-training is beneficial for image-text retrieval benchmarks, but the effects are minor (Table~\ref{tab:ablation_pretr}).

\begin{table}[h]
\centering
\small
\tablestyle{8pt}{1.1}
\begin{tabular}{l|cc|cc}
% \toprule
& 
\multicolumn{2}{c|}{MS COCO} & \multicolumn{2}{c}{Flickr30K} \\
 pre-training? & I2T & T2I &I2T &T2I \\ 
\hline
 No  & 54.2    & \bf{38.9}  & 80.5    & 66.5    \\
 Yes  & \bf{54.9}   & 38.9  & \bf{82.4}   & \bf{67.3}   \\
\hline
% \bottomrule
\end{tabular}
\vspace{2mm}
\caption{\textbf{Effects of language-side pre-training.} Zero-shot image-text retrieval results on COCO and Flickr30K benchmarks, R@1 shown. Language pre-training provides small but consistent improvements.}
\label{tab:ablation_pretr}
\vspace{-2mm}
\end{table}

\textbf{Projections and attention pooling.}
Table~\ref{tab:ablation_proj} experiments with additional projections and attention pooling. We find that they are not needed for joint learning, indicating that the text decoder can be fully shared between tasks, which is a surprising finding. We only train for 100K iterations here.

\begin{table}%[t]
\centering
\small
\tablestyle{5.5pt}{1.1}
\begin{tabular}{l|ccc|ccc}
% \toprule
%  & \multicolumn{6}{c|}{Flickr30K (1K test set)}  \\
%  & \multicolumn{3}{c}{\underline{{\white{-------}}image-to-text{\white{-------}}}} & \multicolumn{3}{c|}{\underline{{\white{-------}}text-to-image{\white{-------}}}} \\ %&/ 
 & \multicolumn{3}{c|}{image-to-text} & \multicolumn{3}{c}{text-to-image} \\ %&/ 
 \hline
Att. Pool/ Proj       & R@1  & R@5  & R@10 & R@1  & R@5   & R@10      \\
%\shline
\hline
N/N &53.32 &80.57 &86.91 &39.86 &67.7 &77.46 \\
Y/N &52.44 &78.71 &86.33 &39.3 &66.62 &76.91 \\
N/Y &51.37 &78.81 &86.13 &39.49 &66.97 &77.05 \\
Y/Y &50.78 &76.95 &85.74 &38.59 &66.54 &76.07 \\
\hline
% \bottomrule
\end{tabular}
\vspace{2mm}
\caption{\textbf{Projection and attention pooling.} Our conclusions are that they are not needed to realize joint training. Retrieval performance on Flickr30K is shown.}
\label{tab:ablation_proj}
\vspace{-2mm}
\end{table}

\textbf{Bi-directional masking is important for contrastive learning.}
Since here we are using a single language decoder to perform tasks of various characteristics e.g. retrieval and VQA, it is important to understand how to construct the language decoder for such tasks. We evaluated the effect of bi-directional or causal masking on a contrastive learning task on the Flickr dataset. We note that with these tasks, our model has to accommodate retrieval tasks which tend to process data with short text, whereas generative tasks tend to generate longer-text queries. %At the same time retrieval tasks have the opportunity to consider the full query. 
We find that bi-directional masking benefits the contrastive learning much more than causal (Table~\ref{tab:ablation_bidir}), which is not surprising as the text associated with retrieval tasks is fully available before final feature representation is generated. 

\begin{table}[h]
\centering
\small
\tablestyle{8pt}{1.1}
\begin{tabular}{l|cc|cc}
% \toprule
& 
\multicolumn{2}{c|}{MS COCO} & \multicolumn{2}{c}{Flickr30K} \\
 Masking & I2T & T2I &I2T &T2I \\ 
\hline
 Causal  & 53.0    & 37.5  & 78.1  & 62.5    \\
 Bi-directional  & \bf{54.9}   & \bf{38.8}  & \bf{82.4}   & \bf{67.3}   \\
\hline
\end{tabular}
\vspace{2mm}
\caption{\textbf{Bi-directional masking.} Contrastive learning clearly benefits from Bi-directional masking in the decoder language model.}
\label{tab:ablation_bidir}
\vspace{-2mm}
\end{table}

\subsection{Total Train Compute Usage}
We present the total compute used to train \ours models in comparison with other foundational models by using the approximation technique in~\cite{gpt3}. Some compute estimates are taken from~\cite{pali} e.g. PaLI~\citep{pali}, Flamingo~\citep{flamingo}, GIT-2~\citep{wang2022git} while others are estimated based on the paper details e.g. CoCa~\citep{yu2022coca}. The total train compute usage of \ours is significantly lower than existing foundational models, for example, 3.4x cheaper than PaLI, 5.5x than CoCa, 10.3x than Flamingo, or 41.2x than GIT-2. Notably, \ours is trained from scratch while some existing methods e.g. PaLI and Flamingo rely on pretrained image encoders either from contrastive learning or image classification.

% \begin{table}[h]
% \centering
% \small
% \tablestyle{8pt}{1.1}
% \begin{tabular}{l|cccc|c}
% Method & CoCa & PaLI & Flamingo & GIT-2 & \ours(Ours) \\ 
% \hline
% PF-days & 743 & 453 & 1381 & 5513 & 134    \\
% Ratio to ours & 5.5$\times$ & 3.4$\times$ & 10.3$\times$ & 41.2$\times$ & 1$\times$  \\
% \hline
% % \bottomrule
% \end{tabular}
% \caption{\textbf{Total Compute Comparison.} We compare the total train compute use with existing foundational models. \ours is significantly cheaper to train than existing foundational models without using any pretrained models.}
% \label{tab:ablation_pretr}
% \vspace{-2mm}
% \end{table}

\end{document}